\documentclass[sigconf]{acmart}

\copyrightyear{2025}
\acmYear{2025}
\setcopyright{acmlicensed}\acmConference[MM '25]{Proceedings of the 33rd ACM International Conference on Multimedia}{October 27--31, 2025}{Dublin, Ireland}
\acmBooktitle{Proceedings of the 33rd ACM International Conference on Multimedia (MM '25), October 27--31, 2025, Dublin, Ireland}
\acmDOI{10.1145/3746027.3754885}
\acmISBN{979-8-4007-2035-2/2025/10}

\usepackage[utf8]{inputenc} 
\usepackage[T1]{fontenc}    
\usepackage{url}            
\usepackage{booktabs}       
\usepackage{amsfonts}       
\usepackage{nicefrac}       
\usepackage{microtype}      
\usepackage{amsthm}
\usepackage{amsmath}
\usepackage{graphicx}

\usepackage{amssymb}
\usepackage{bbm}
\usepackage{diagbox}
\usepackage{colortbl}
\usepackage{multirow}
\usepackage{pifont} 
\usepackage{bm}
\usepackage{subcaption}
\usepackage{color,xcolor}
\usepackage[font=small,labelfont=bf]{caption}  
\usepackage{makecell}
\usepackage{tabulary}
\definecolor{demphcolor}{RGB}{144,144,144}

\definecolor{mygray}{gray}{0.4}
\usepackage{colortbl}
\usepackage{enumitem}
\usepackage{multirow}
\usepackage{threeparttable}
\usepackage{setspace}
\usepackage[export]{adjustbox}

\usepackage{makecell}

\usepackage{xcolor} 
\definecolor{snowblue}{rgb}{0.65, 0.80, 0.98} 

\usepackage{newfloat}
\usepackage{listings}

\AtBeginDocument{%
  \providecommand\BibTeX{{%
    Bib\TeX}}}

\AtBeginDocument{%
  \providecommand\BibTeX{{%
    \normalfont B\kern-0.5em{\scshape i\kern-0.25em b}\kern-0.8em\TeX}}}

\begin{document}

\title{FineQuest: Adaptive Knowledge-Assisted Sports Video Understanding via Agent-of-Thoughts Reasoning}

\author{Haodong Chen}
\orcid{0009-0009-1666-2037}
\affiliation{%
  \institution{School of Automation, Northwestern Polytechnical University}
  \city{}
  \country{}}
\email{haroldchen328@gmail.com}

\author{Haojian Huang}
\orcid{0000-0002-0661-712X}
\affiliation{%
  \institution{The University of Hong Kong}
  \city{Hong Kong}
  \country{China}}
\email{haojianhuang927@gmail.com}

\author{Xinxiang Yin}
\orcid{0009-0009-3021-6439}
\affiliation{%
  \institution{School of Software, Northwestern Polytechnical University}
  \city{Xi'an City}
  \country{China}}
\email{yinxinxiang@mail.nwpu.edu.cn}

\author{Dian Shao}
\orcid{0000-0002-0862-9941}
\affiliation{%
  \institution{Unmanned System Research Institute, Northwestern Polytechnical University}
  \city{Xi'an City}
  \country{China}}
\email{shaodian@nwpu.edu.cn}
\authornote{Corresponding author}

\renewcommand{\shortauthors}{Haodong Chen, Haojian Huang, Xinxiang Yin, and Dian Shao}


\begin{abstract}
Video Question Answering (VideoQA) based on Large Language Models (LLMs) has shown potential in general video understanding but faces significant challenges when applied to the inherently complex domain of sports videos. In this work, we propose \textbf{FineQuest}, the first training-free framework that leverages dual-mode reasoning inspired by cognitive science: i) \textit{\textbf{Reactive Reasoning}} for straightforward sports queries and ii) \textit{\textbf{Deliberative Reasoning}} for more complex ones. To bridge the knowledge gap between general-purpose models and domain-specific sports understanding, FineQuest incorporates \textbf{SSGraph}, a multimodal sports knowledge scene graph spanning nine sports, which encodes both visual instances and domain-specific terminology to enhance reasoning accuracy. Furthermore, we introduce two new sports VideoQA benchmarks, \textbf{Gym-QA} and \textbf{Diving-QA}, derived from the FineGym and FineDiving datasets, enabling diverse and comprehensive evaluation. FineQuest achieves state-of-the-art performance on these benchmarks as well as the existing SPORTU dataset, while maintains strong general VideoQA capabilities.

\end{abstract}




\maketitle

\begin{figure}[!t]
    \centering
   \includegraphics[width=1\linewidth]{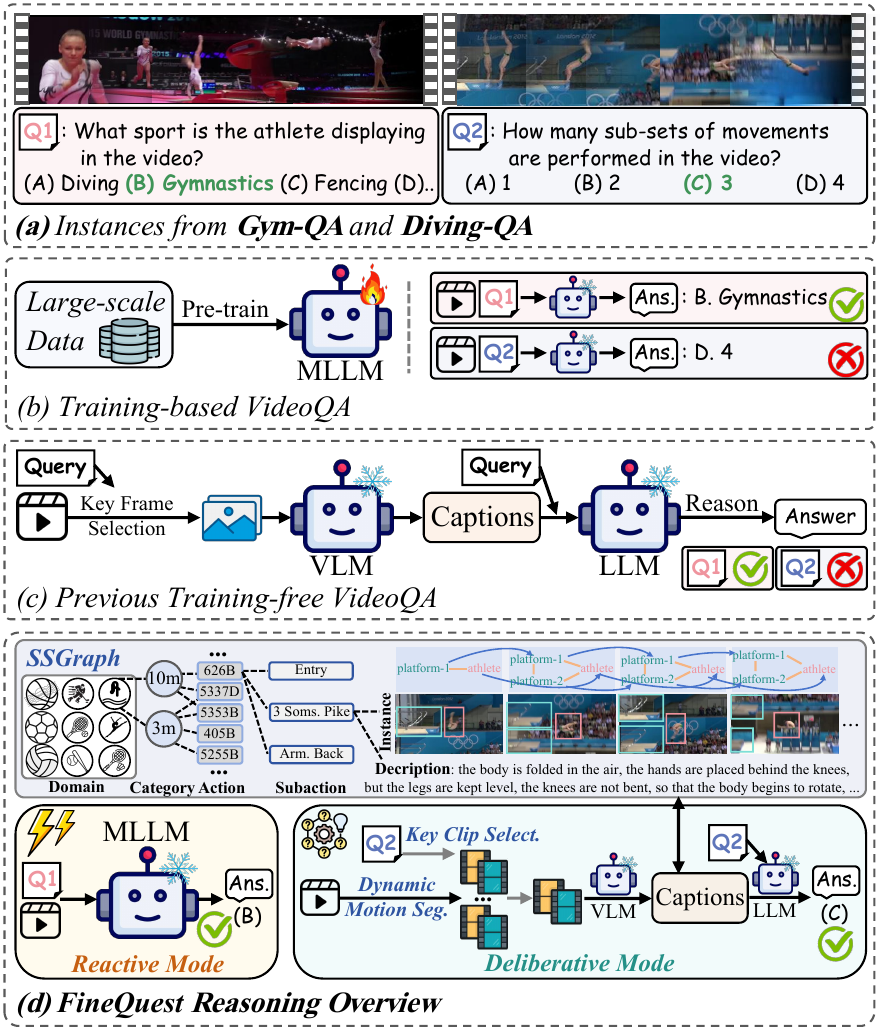}
   \vspace{-2.2em}
    \captionof{figure}{\label{fig:teaser}
    (\textbf{\textit{a}}) Constructed \textbf{Gym-QA} and \textbf{Diving-QA} examples (Q1: easy, Q2: hard). (\textbf{\textit{b}}) Training-based VideoQA requires large annotated datasets for training but struggles to answer complex questions like Q2 (response from GPT-4V \cite{openai2024hello}). (\textbf{\textit{c}}) Existing training-free VideoQA uses a VLM (captioner) and an LLM (reasoner) but fails on complex sports queries like Q2 (response from VideoTree \cite{wang2024videotree}). (\textbf{\textit{d}}) \textit{FineQuest} combines Reactive Mode for simple queries and Deliberative Mode with Dynamic Motion Segmentation, Key Clip Selection, and SSGraph to address complex questions progressively.}
    \vspace{-1.6em}
\end{figure}

\vspace{-1em}
\section{Introduction}
\label{sec:intro}

Sports events captivate audiences worldwide, offering moments of intense action, strategic play, and athletic excellence. With the increasing demand for personalized and interactive viewing experiences \cite{yang2024sports}, Video Question Answering (VideoQA) has emerged as a promising technology to enable real-time insights and user-driven exploration of sports content. 


However, understanding sports videos presents unique challenges that go beyond those encountered in general video understanding tasks \cite{yu2024self, ma2024beyond, lin2023mm, wang2022language, wang2024videotree, chen2024finecliper, song2024moviechat, huang2025sefar}: \ding{182}~\textbf{Complex and rapid actions} require fine-grained motion understanding. \ding{183}~\textbf{Multi-camera perspectives and multi-object interactions} demand sophisticated spatiotemporal reasoning. \ding{184}~\textbf{Strong contextual dependencies} necessitate domain-specific knowledge for accurate event interpretation. \ding{185}~\textbf{Data scarcity} makes large-scale, high-quality sports video annotation prohibitively expensive, limiting the effectiveness of traditional tuning-based approaches (see Fig.\textcolor{magenta}{~\ref{fig:teaser}}(b)), which motivates the exploration of \textit{training-free VideoQA} methods.

Existing training-free VideoQA methods \cite{yu2024self, wang2024videotree, park2024too}, while promising, face significant shortcomings when applied to sports videos (see Fig.\textcolor{magenta}{~\ref{fig:teaser}}(c)) \cite{shao2020finegym, xu2022finediving}. Most rely on scene-based clustering \cite{hartigan1979algorithm}, which struggles to handle the frequent camera switches inherent in sports broadcasts. Additionally, they often select a single frame or a small set of frames per cluster for captioning, which fails to capture the continuous and complex nature of sports actions. Furthermore, general-purpose vision-language models (VLMs) \cite{li2023llama, huang2025vistadpo, liu2024llava, jin2024chat, lin2023video, maaz2023video, ma2023vista} lack the specialized knowledge required to map visual cues to professional sports terminology (\textit{e.g.}, "what `626B' refers to?"), leading to poor performance on domain-specific tasks.

Adding to these challenges is the diverse complexity of sports VideoQA queries (see Fig.\textcolor{magenta}{~\ref{fig:teaser}}(a)). 
For instance, a simple queries like "What sport is this?" rely on basic pattern recognition, while complex ones, such as "How many sub-sets of movements are performed?", demand multi-step reasoning involving segment identification, rule comprehension, and performance analysis. This highlights the limitations of one-size-fits-all reasoning approaches \cite{li2024mvbench, yu2024self, lin2023mm, wang2022language}, underscoring the need for a dynamic framework tailored to varying complexity and domain-specific requirements.

To address these challenges, we propose \textbf{FineQuest}, a novel training-free sports VideoQA framework that incorporates Agent Chain-of-Thought (CoT) planning \cite{sanwal2025layered, shi2024unlocking} as shown in Fig.\textcolor{magenta}{~\ref{fig:teaser}}(d). Inspired by dual-process theories in cognitive science, FineQuest operates in two complementary modes of reasoning:
\vspace{-0.4em}
\begin{itemize}[leftmargin=*]
    \item[\ding{238}] \textbf{Reactive Reasoning}: Handles simple, low-complexity queries with a single-step process, leveraging pre-trained (M)LLMs for rapid responses.
    \item[\ding{238}] \textbf{Deliberative Reasoning}: Tackles complex, high-context queries using multi-step reasoning guided by Agent CoT planning. For example, answering Q2 in Fig.\textcolor{magenta}{~\ref{fig:teaser}} involves: \ding{172} identifying relevant video segments via \textit{Dynamic Motion Segmentation}, \ding{173} selecting key clips based on the query using \textit{Key Clip Selection}, and \ding{174} leveraging domain-specific knowledge for \textit{Fine-grained Matching}. This mode systematically decomposes the query into sub-tasks and reasoning through multiple stages.
\end{itemize}
\vspace{-0.4em}
This dual-mode reasoning framework ensures both effectiveness and robustness, allowing FineQuest to handle diverse queries with precision and adaptability. However, the lack of domain-specific knowledge integration is a major limitation of existing VideoQA systems. To bridge this gap, we introduce \textbf{SSGraph} (see top of Fig.\textcolor{magenta}{~\ref{fig:teaser}}(d)), the first multimodal sports knowledge scene graph. SSGraph spans \textbf{\textit{nine}} sports (\textit{i.e.}, gymnastics, diving, basketball, soccer, ice hockey, tennis, baseball, badminton, and volleyball) and integrates: \textit{\textbf{(i)}}~\textbf{Visual instances}: Representing key objects, actions, and attributes from sports videos. \textit{\textbf{(ii)}}~\textbf{Domain-specific terminology}: Encoding rules, scoring criteria, and contextual dependencies unique to each sport. SSGraph serves as a robust knowledge supplementation for reasoning, enabling FineQuest to answer questions that require a precise, context-aware understanding of sports scenes.

Additionally, to address the limitations of existing sports QA benchmarks \cite{xia2024sportu, li2024sports}, which often feature limited query diversity \cite{li2024sports} and lack coverage of varying intensities \cite{xia2024sportu}, we build upon two high-quality action understanding datasets, FineGym \cite{shao2020finegym} and FineDiving \cite{xu2022finediving}, to create Gym-QA and Diving-QA, for more diverse and comprehensive evaluation in the field as shown in Fig.\textcolor{magenta}{~\ref{fig:teaser}}(a). While the annotation process still requires specialized domain knowledge and cannot rely as extensively on (M)LLMs as recent works \cite{mangalam2024egoschema, ataallah2024infinibench}, we introduce rule-based filtering procedures and leverage the advanced capabilities of LLMs to significantly reduce the burden on human annotators. Briefly put, our contributions can be summarized as:
\vspace{-0.4em}
\begin{itemize}[leftmargin=*]
    \item We propose \textbf{FineQuest}, the first training-free framework designed to address the challenging task of sports VideoQA, integrating agent CoT reasoning with two modes: \textit{Reactive Reasoning} for efficient performance on general or simple queries. \textit{Deliberative Reasoning} for multi-step reasoning on complex queries.
    \item To meet the specialized domain knowledge requirements for understanding sports videos, we develop SSGraph, the first multimodal sports knowledge scene graph, integrating domain-specific terminology and visual instances to enhance reasoning accuracy.
    \item Extensive experiments demonstrate FineQuest achieves state-of-the-art results on our newly built sports VideoQA benchmarks, namely Gym-QA and Diving-QA, as well as existing SPORTU \cite{xia2024sportu} benchmark, while maintaining strong general QA performance.
\end{itemize}


\section{Related Work}
\label{sec:related}

\vspace{-0.1em}
\subsubsection*{\textbf{Sports Video Understanding}}
has evolved significantly, from early work on action recognition \cite{wang2018temporal, shao2020intra, carreira2017quo, shao2018find} and gameplay phase categorization \cite{cabado2022real} to broader tasks like player detection \cite{vandeghen2022semi}, tracking \cite{goel2023humans}, action spotting \cite{vanderplaetse2020improved}, tactical analysis \cite{suzuki2019team}, and generation \cite{chen2025hierarchical, shao2025finephys, chen2025temporal}. Datasets like FineGym \cite{shao2020finegym}, FineDiving \cite{xu2022finediving}, MultiSports \cite{li2022multi}, and soccer-specific datasets such as SoccerNet \cite{giancola2018soccernet}, SoccerDB \cite{jiang2020soccerdb}, SoccerTrack \cite{scott2022soccertrack} and UniSoccer \cite{rao2024towards} have driven these advancements.

Despite these advancements, sports VideoQA—a task that enables interactive video understanding—remains underexplored. Currently, only a few benchmarks exist for this purpose \cite{li2024sports, xia2024sportu}. Specifically, Sports-QA \cite{li2024sports}, built on \cite{shao2020finegym, Li_2021_ICCV}, offers limited query depth and diversity, as it primarily extends action labels into simple questions. While SPORTU \cite{xia2024sportu} 
considers query difficulty more thoroughly, all video clips in SPORTU are in slow motion, which limits its diversity and applicability to real-world scenarios. 
To address this, we expand FineGym \cite{shao2020finegym} and FineDiving \cite{xu2022finediving} to create Gym-QA and Diving-QA, introducing diverse query difficulties and supporting high-speed motion analysis, thereby filling key gaps in sports VideoQA research.

\begin{figure*}[!t]
    \centering
    \vspace{-0.1em}
   \includegraphics[width=1\linewidth]{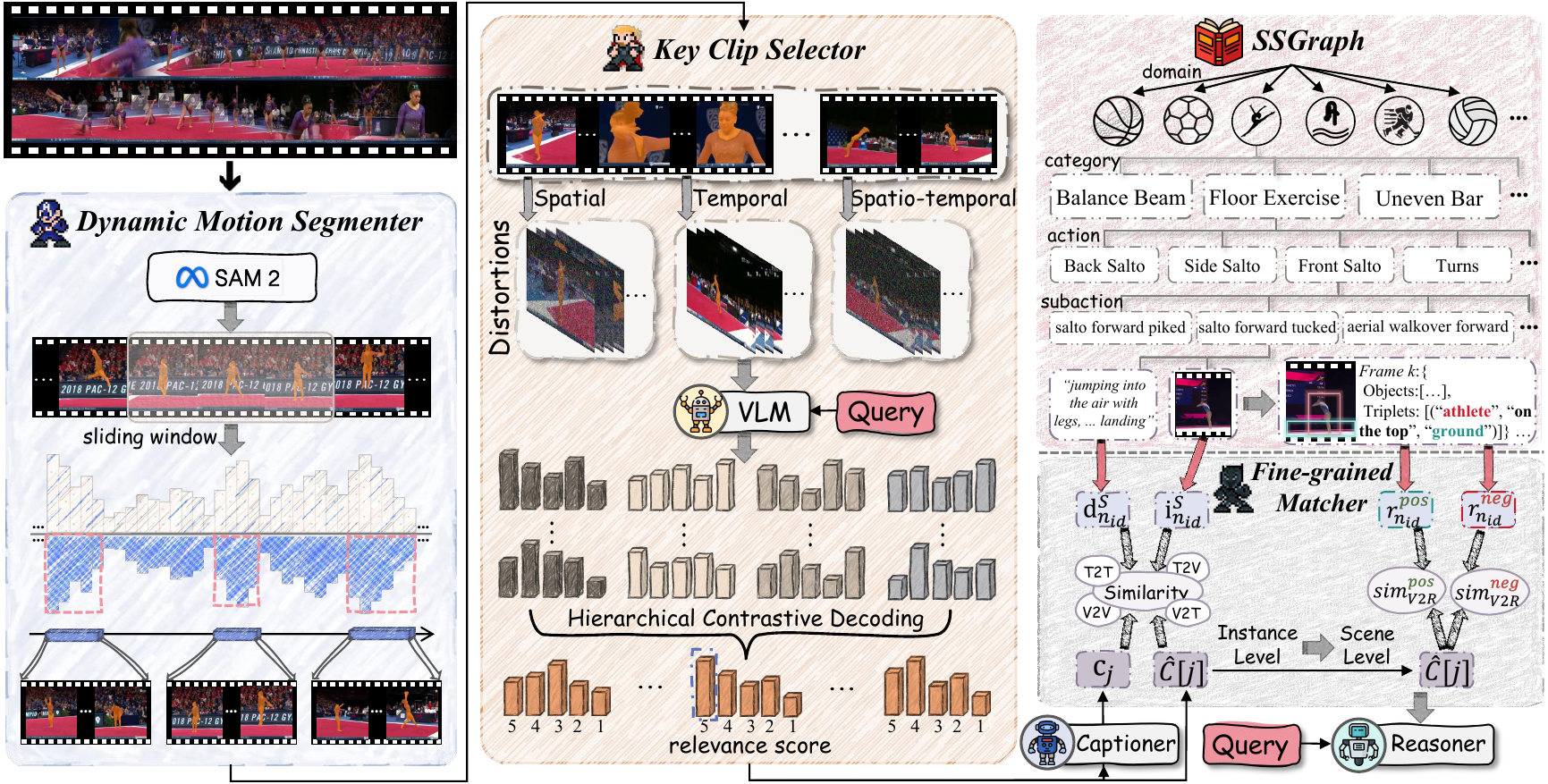}
   \vspace{-2.2em}
    \captionof{figure}{\label{fig:framework}
    Deliberative Mode of \textit{FineQuest}: Activated via a system switch for complex sports VideoQA tasks, it includes (1) Dynamic Motion Segmenter for adaptive sub-action segmentation, (2) Key Clip Selector for hierarchical contrastive decoding to identify query-relevant clips, and (3) SSGraph-based Fine-grained Matcher for precise query matching using structured instance-to-scene knowledge. }
    \vspace{-0.8em}
\end{figure*}

\vspace{-0.4em}
\subsubsection*{\textbf{Video Question Answering}}
generates open-ended textual responses based on video content and queries. Early work focused on short-term videos (5-30 seconds) using pretraining-finetuning \cite{yu2024self}, zero-shot \cite{lin2023mm}, and few-shot learning \cite{wang2022language}. Recent advancements have expanded its scope to longer formats like movies and vlogs \cite{wang2024videotree, zhang2024flash, song2024moviechat}, supported by benchmarks \cite{mangalam2024egoschema, xiao2021next, fang2024mmbench, fu2024video} such as InfiniBench \cite{ataallah2024infinibench}. Concurrently, the emergence of LLMs has further inspired new approaches to VideoQA \cite{zhang2023simple, wang2023vamos, wang2024videoagent, kahatapitiya2024language, sun2024visual}. A prevalent strategy involves using VLMs to convert frame-level visual information into natural language, which is then processed by LLMs to generate predictions.

Despite these advancements, existing models often neglect domain-specific challenges. Sports VideoQA, for instance, demands fine-grained motion analysis, multi-object interactions, and domain expertise. To address these gaps, we propose FineQuest, the first training-free framework tailored for sports VideoQA. FineQuest integrates agent Chain-of-Thought (CoT) reasoning with two complementary modes: \textit{Reactive Reasoning} for handling general and simple sports queries efficiently; \textit{Deliberative Reasoning} for multi-step reasoning on complex, context-rich sports queries.

\vspace{-0.4em}
\subsubsection*{\textbf{Knowledge Scene Graph.}} Multi-modal knowledge graphs (MMKGs) enhance traditional knowledge bases by integrating visual and textual dimensions into standard triples (head, relation, tail) \cite{ferrada2017imgpedia, liu2019mmkg}. Examples include VisualSem \cite{alberts2020visualsem}, which enriches nodes with images and glosses, and PKG \cite{li2022multi}, linking Chinese poetry with visuals.

However, MMKGs are limited to static knowledge and struggle to represent dynamic, context-sensitive scenarios like sports events. Knowledge Scene Graphs (KSGs) address this by incorporating spatiotemporal and contextual data to model interactions within events \cite{huang2024structure, fei2024video}. Building on this, we introduce SSGraph, the first sports-specific KSG, integrating multi-modal data and domain-specific knowledge to model dynamic interactions in sports. SSGraph bridges MMKGs and KSGs, enabling advanced reasoning for tasks like action recognition, quality assessment, and VideoQA.

\vspace{-0.7em}
\section{Preliminaries}

The integration of components in visual agents $\mathcal{F}$ typically involves three key modules: a visual encoder $\mathcal{E}_V$, a natural language text encoder $\mathcal{E}_T$, and a large language model such as Vicuna~\cite{vicuna2023}. The process begins with the visual agent receiving a video $\mathcal{V}$ and an accompanying textual prompt $\mathcal{Q}$, which could be a query or an instruction. These inputs are then processed into a unified multimodal representation space. Finally, the visual agent generates a textual response $\mathcal{R}$ based on the combined video and textual input. The generation process can be formalized as:
\setlength\abovedisplayskip{4pt}
\setlength\belowdisplayskip{4pt}
\begin{equation}
\mathcal{R}=\mathcal{F}\left[\mathcal{E}_V(\mathcal{V}),\mathcal{E}_L(\mathcal{Q})\right].
\label{eq1}
\end{equation}

\noindent\textbf{Definition} \textit{(System 1 and System 2 Thinking) System 1 and System 2 are two distinct modes of thinking proposed by Nobel Laureate Daniel Kahneman in his book Thinking, Fast and Slow \cite{kahneman2011thinking}.}

\vspace{0.44em}
\noindent\textit{System 1 - Fast Thinking (Reactive Mode): Unconscious, automated thinking processes that are fast, intuitive, and effortless. It is responsible for automatic responses and basic cognitive operations in daily activities. However, it is prone to heuristic biases and errors. Examples include recognizing familiar faces or locating objects in a room.}

\vspace{0.44em}
\noindent\textit{System 2 - Slow Thinking (Deliberative Mode): Conscious, deliberate thinking processes that are slow, effortful, logical, and analytical. It handles complex calculations, reasoning, and decision-making while also monitoring and controlling System 1 processes. Examples include filling out a tax form or analyzing the structure of a sentence.}

\vspace{-0.44em}
\section{SSGraph: Sports Knowledge Scene Graph}

A critical gap in sports video understanding compared to general video understanding is the lack of domain-specific knowledge in existing models. Addressing this through retraining or fine-tuning is challenging due to the diversity and scarcity of high-quality annotated sports data.

To bridge this gap, we propose SSGraph, the first multi-modal sports knowledge scene graph, representing domain-specific knowledge across nine sports: \underline{G}ymnastics, \underline{D}iving, \underline{B}asketball, \underline{S}occer, \underline{I}ce Hockey, \underline{T}ennis, \underline{B}aseball, \underline{B}adminton, and \underline{V}olleyball.
Each sports element in SSGraph is a tuple $\langle n_i, r_j, n_k \rangle$, connecting nodes $n_i$ and $n_k$ via relation $r_j$, capturing complex sports interactions.

\vspace{-0.4em}
\subsubsection*{\textbf{Raw Data Collection.}}
Unlike most MMKGs \cite{alberts2020visualsem, lee2024multimodal, liu2019mmkg} that focus on comprehensive knowledge (\textit{e.g.}, from Wikipedia), sports knowledge is specialized and structured, allowing much easier modification of sports disciplines. We constructed nine sports categories for new benchmarks, Gym-QA and Diving-QA, using datasets like FineGym~\cite{shao2020finegym}, FineDiving~\cite{xu2022finediving}, MultiSports~\cite{li2022multi}, and internet sources.

\vspace{-0.4em}
\subsubsection*{\textbf{Knowledge Processing.}} Specialized sports terminology poses challenges for understanding. We first collected videos corresponding to all sports elements (\textit{i.e.}, the finest-grained sub-actions) from the source datasets. Notably, the selection of videos requires adherence to specific criteria: \ding{182} \textit{high-definition quality}, \ding{183} \textit{minimal camera switching}, and \ding{184} \textit{good lighting conditions}, to better facilitate subsequent steps. Based on this, we meticulously watched each collected video multiple times to provide detailed descriptions of the athletes' movements, serving as the general descriptions for each element action label. During this process, we also relied on external sources (\textit{e.g.}, websites\footnote{https://gymnasticshq.com/}) to ensure that our descriptions are accurate and professional.

\vspace{-0.4em}
\subsubsection*{\textbf{Nodes Connection.}} The structured nature of sports allows hierarchical construction of SSGraph as an MMKG (\textit{e.g.}, Gymnastic → event → set → element), connecting element nodes with video and description nodes.

\vspace{-0.4em}
\subsubsection*{\textbf{Scene Representation.}}
To enhance spatiotemporal modeling, SSGraph is extended into a knowledge scene graph. A spatiotemporal scene graph (STSG) comprises sequences of individual scene graphs (SGs) for each video frame, capturing core semantic representations. Each SG contains "\textit{subject}"-"\textit{predicate}"-"\textit{object}" triplets, where "\textit{subject}" and "\textit{object}" represent two visual proposals (RoIs), and "\textit{predicate}" defines the relationship between them, filtering less-informative details and aiding video content understanding. Each SG at frame $t$ is $\mathcal{G}_t=(I_t;E_t)$, with sub-nodes $v{t,i}$ connected by edges $e_{t,i,j}$. Temporal coreference edges $e_{t-1\rightarrow t}$ link objects across frames, ensuring dynamic interaction tracking.

\section{Methodology}

In this section, we present \textit{FineQuest}, a novel training-free framework designed to address the challenging task of sports VideoQA. FineQuest features a dynamic system switch mechanism that enables rapid responses to simple sports questions or general questions (Reactive Mode) and accommodates deliberate reasoning for intricate questions (Deliberative Mode) as shown in Fig.\textcolor{magenta}{~\ref{fig:framework}}.

\vspace{-0.32em}
\subsection{Reactive Mode with System Switch}
\vspace{-0.1em}

Current works on visual agents for VideoQA primarily rely on visual question-answering datasets, which directly provide answers after inquiry as Eq.~(\textcolor{magenta}{\ref{eq1}}). However, answering questions in this manner can compromise the reliability of responses \cite{sun2024visual}. Specifically, visual agents often hallucinate when handling sports-related questions that require deliberate reasoning, detailed visual understanding, and external domain-specific knowledge. To mitigate hallucination and enhance the reliability of the framework, we propose a system switch trigger mechanism that determines when deliberate reasoning is required. For a given question $\mathcal{Q}$ and video $\mathcal{V}$, we define an MLLM as a Reactive Reason Agent $\mathcal{F}_{\text{React}}(\cdot)$, equipped with the reasoning process to analyze the difficulty of the question:
\setlength\abovedisplayskip{3.6pt}
\setlength\belowdisplayskip{3.6pt}
\begin{equation}
\mathcal{R}_{\text{React}}=\mathcal{F}_{\text{React}}\left[\mathcal{E}_V(\mathcal{V}),\mathcal{E}_L(\mathcal{Q})\right].
\label{eq2}
\end{equation}

Specifically, the reactive reason agent follows a structured reasoning process: \ding{182} \textbf{\textit{Query Difficulty Analysis}}: The agent first evaluates the difficulty of the question $\mathcal{Q}$ based on the following dimensions: (\textbf{\textit{a}}) \textit{Question-Video Relevance}: Can the video directly answer the question? (\textbf{\textit{b}}) \textit{Question Type}: Static (\textit{e.g.}, action matching) \textit{vs.} dynamic (\textit{e.g.}, event inference). (\textbf{\textit{c}}) \textit{Reasoning Requirements}: Single-step \textit{vs.} multi-step inference. (\textbf{\textit{d}}) \textit{External Knowledge Dependency}: Need for domain-specific knowledge. \ding{183} \textbf{\textit{Decision Making}}: Based on the difficulty analysis, the Reactive Reason Agent selects one of the following two actions: (\textit{\textbf{i}}) \textit{Direct Answer Generation}: If the question is determined to be simple, the response $\mathcal{R}_{\text{React}}$ includes both the answer and the reasoning basis for the decision. (\textit{\textbf{ii}}) \textit{System Switch}: If the question is deemed complex, the agent triggers a system switch, delegating the task to the Deliberative Reason Agent for advanced reasoning. The response $\mathcal{R}_{\text{React}}$ includes a "switch" command and an explanation of why the switch is necessary.

\vspace{-0.36em}
\subsection{Deliberative Mode}
\vspace{-0.1em}

Previous training-free VideoQA paradigms \cite{wang2024videotree, park2024too, zhang2023simple} typically employ clustering algorithms (\textit{e.g.}, k-means \cite{hartigan1979algorithm} in VideoTree \cite{wang2024videotree}) to identify scene transitions and segment the video $\mathcal{V}$ into clips. The keyframes are selected based on the query $\mathcal{Q}$, and a VLM acts as a captioner $\mathcal{P}_{Caption}(\cdot)$ to generate captions for these keyframes. These captions, along with $\mathcal{Q}$, are provided as input to an LLM, which acts as a reasoner $\mathcal{P}_{Reason}(\cdot)$ to predict the answer.

While effective for general videos, this paradigm faces challenges with sports videos due to: (\textit{\textbf{i}}) Sports videos often feature single, continuous scenes with complex backgrounds, making scene-based segmentation ineffective. (\textbf{\textit{ii}}) Frequent shot transitions in sports videos are often misclassified as scene changes by clustering algorithms. (\textbf{\textit{iii}}) Keyframes fail to capture the full semantics of complex sports actions. To address these limitations, we further propose the Deliberative Reason Agent $\mathcal{F}_{\text{Deliberate}}(\cdot)$ to decompose the difficulty of the input query step by step.

\vspace{-0.4em}
\subsubsection*{\textbf{Dynamic Motion Segmenter.}} To mitigate the ambiguity caused by complex backgrounds, we adopt the powerful SAM 2 \cite{ravi2024sam2} to highlight athletes in sports videos. This ensures the preservation of essential background context, enabling accurate captioning by the subsequent $\mathcal{P}_{Caption}(\cdot)$.

Given the lack of well-annotated sports video datasets \cite{shao2020finegym}, relying on pre-training with existing timestamp annotations may result in poor generalization. To address this, we propose the Dynamic Motion Segmenter $\mathcal{S}_{segment}(\cdot)$, which adaptively segments sub-actions without requiring predefined timestamps. This method segments sub-actions based on motion intensity, which naturally fluctuates during athletic performance. Specifically, athletes exhibit varying motion intensities, with brief pauses between action sets \cite{zhao2017temporal}. To capture these dynamics, we employ a sliding window to adjust the segmentation threshold based on local motion magnitude. This adaptive process aligns with natural human kinematics, enabling precise generation of sub-action proposals (details can be found in Algorithm~\textcolor{magenta}{1} in \textbf{Appendix}~\S\textcolor{magenta}{B}):
\begin{equation}
    Proposals = \mathcal{S}_{segment}(\hat{\mathcal{V}})
\end{equation}


\vspace{-1em}
\subsubsection*{\textbf{Key Clip Selector.}}

Given the segmented clips $\mathcal{C}[i]$ for $i \in \{1, K\}$, the query $\mathcal{Q}$ often pertains to only a small portion of the video. While prior works utilize VLMs (\textit{e.g.}, CLIP in LVNet \cite{park2024too} or BLIP-2 in SeViLA \cite{yu2024self}) to directly measure frame-query similarity, this approach is prone to hallucinations \cite{tong2024eyes}, caused by biases in training data \cite{goyal2017making} and over-reliance on language priors \cite{li2023evaluating}.

To address this, we propose the Key Clip Selector $\mathcal{S}_{select}(\cdot)$, which employs \textit{contrastive decoding} \cite{li2022contrastive, leng2024mitigating} to enhance relevance scoring by contrasting outputs from original and distorted inputs. Specifically, given a VLM-based selector parameterized by $\theta$, the model processes $\mathcal{Q}$ and a visual input $\mathcal{C}[i]$, generating a response $y$ sampled from the probability distribution:
\begin{equation}
    y_t \sim p_\theta\left(y_t\mid \mathcal{C}[i],\mathcal{Q'},y_{<t}\right),
\end{equation}
where $\mathcal{Q'}$ augments the query $\mathcal{Q}$ with a selection prompt $p_s$, and $y_{<t}$ denotes the sequence of tokens generated up to time step $t-1$. To mitigate hallucinations, we introduce a distorted version $\mathcal{C}_{Dis}[i]$ of $\mathcal{C}[i]$, producing two output distributions. Their differences are used to compute a contrastive probability distribution:
\begin{equation}
    \begin{aligned}&p(y\mid \mathcal{C}[i],\mathcal{C}_{Dis}[i],\mathcal{Q'})=\operatorname{softmax}[(1+\alpha)\\&\operatorname{logit}_\theta\left(y\mid \mathcal{C}[i],\mathcal{Q'}\right)-\alpha\operatorname{logit}_\theta\left(y\mid \mathcal{C}_{Dis}[i],\mathcal{Q'}\right)],\end{aligned}
    \label{eq2}
\end{equation}
where $\alpha$ controls the amplification of contrastive differences.

Unlike prior works focused on images \cite{leng2024mitigating}, our input consists of video clips, requiring distortions that preserve semantic integrity across spatial and temporal dimensions. We propose three types of distortions: \ding{182}~\textit{\textbf{Spatial}}: Adding Gaussian noise to each frame. \ding{183}~\textit{\textbf{Temporal}}: Applying temporal warping\footnote{Distortion on the temporal duration of each frame while keeping the order unchanged} \cite{xing2023svformer} to adjust frame durations while maintaining order. \ding{184}~\textit{\textbf{Spatio-Temporal}}: Combining spatial and temporal distortions. These methods avoid issues with approaches like CutMix \cite{yun2019cutmix}, which can obscure fine-grained actions, or frame shuffling, which disrupts temporal coherence \cite{huang2025sefar}. For each clip $\mathcal{C}[i]$, the distorted versions are denoted as $\mathcal{C}_{Spa}[i]$, $\mathcal{C}_{Tem}[i]$, and $\mathcal{C}_{ST}[i]$, with corresponding contrastive weights $\alpha_S$, $\alpha_T$, and $\alpha_{ST}$, where $\alpha = \alpha_S + \alpha_T + \alpha_{ST}$. The contrastive probability distribution is then reformulated as:
\begin{equation}
    \begin{aligned}y_{t}\sim&\operatorname{softmax}[(1+\alpha)\operatorname{logit}_\theta\left(y\mid \mathcal{C}[i],\mathcal{Q'}\right)\\&-\alpha_S\operatorname{logit}_\theta\left(y\mid \mathcal{C}_{Spa}[i],\mathcal{Q'}\right)-\alpha_T\operatorname{logit}_\theta\left(y\mid \mathcal{C}_{Tem}[i],\mathcal{Q'}\right)\\&-\alpha_{ST}\operatorname{logit}_\theta\left(y\mid \mathcal{C}_{ST}[i],\mathcal{Q'}\right)].\end{aligned}
    \label{eq3}
\end{equation}

The clips $\hat{\mathcal{C}}[j]$ with the top $N_1$ relevance scores are selected, and adjacent clips are merged. The selected clips are then processed by the VLM-based captioner $\mathcal{P}_{Caption}(\cdot)$ to generate captions:
\begin{equation}
    \mathbf{C} = \mathcal{P}_{Caption}(\mathcal{E}_{V/L}(\mathcal{S}_{select}(\mathcal{C}, \mathcal{Q})))
\end{equation}
where $\mathbf{C}$ represents the textual descriptions of the key clips.

\vspace{-0.4em}
\subsubsection*{\textbf{Fine-grained Matcher.}}

The generated captions $\mathbf{C} = {\mathbf{c}_1, \mathbf{c}_2, ..., \mathbf{c}_{K'}}$ describe selected clips $\hat{\mathcal{C}}[j]$ for $j \in \{1, K'\}$ in general natural language. However, such descriptions often lack the specificity required for answering sports-related queries due to the limited domain knowledge of existing LMs.

To address this limitation, we propose the Fine-grained Matcher $\mathcal{S}_{match}(\cdot)$, which matches domain knowledge from our constructed SSGraph $\mathcal{G}_{SS}$ through a two-level fine-grained matching process:

\ding{182} \textbf{Cross-modal Instance Level.} At this level, we align captions and video clips with specific visual and textual elements, such as objects, actions, and attributes, within SSGraph. For each caption $\mathbf{c}_j$ and its corresponding clip $\hat{\mathcal{C}}[j]$, we compute similarities with the textual description $\mathbf{d}_n^S$ and visual instance $\mathbf{i}_n^S$ of an action node $n_{id}$ in SSGraph, where $S \in \{G, D, B_1, S, I, T, B_2, B_3, V\}$ denotes the sports category. Using Long-CLIP \cite{zhang2024longclip}, which handles multi-modal data and long token sequences, the similarities are calculated via cosine similarity:
\setlength\abovedisplayskip{3.2pt}
\setlength\belowdisplayskip{3.2pt}
\begin{equation}
    sim_{T2T}=\frac{\mathbf{c}_j\cdot\mathbf{d}_{n_{id}}^S}{\left\|\mathbf{c}_j\right\|\left\|\mathbf{d}_{n_{id}}^S\right\|},\quad sim_{V2V}=\frac{\hat{\mathcal{C}}[j]\cdot\mathbf{i}_{n_{id}}^S}{\left\|\hat{\mathcal{C}}[j]\right\|\left\|\mathbf{i}_{n_{id}}^S\right\|},
\end{equation}
where $sim_{T2T}$ and $sim_{V2V}$ represent text-to-text and visual-to-visual matching, respectively. To mitigate ambiguity caused by minimal differences between fine-grained movements, we select the top five sub-actions from SSGraph with the highest similarity to each caption $\mathbf{c}_j$ and clip $\hat{\mathcal{C}}[j]$. These are further matched across modalities using text-to-visual (T2V) and visual-to-text (V2T) processes, ensuring accurate alignment between captions and video clips. However, when textual or visual features are overly detailed, models tend to focus excessively on subjects, leading to hallucinations regarding relationships and interactions between subjects \cite{an2024agla, bai2024hallucination}. This motivates reasoning at a higher relational level.

\begingroup
\setlength{\tabcolsep}{2pt}
\begin{table*}[t]
\fontsize{8.5}{7}\selectfont 
\setlength{\tabcolsep}{1.mm}
  \centering
  \caption{Main results on Sports VideoQA benchmarks. Bold values indicate the best performance and $\Delta$ denotes the corresponding improvement percentages over baselines (\textit{i.e.}, LLaVA-Next-Video and Video-LLaVA). $\uparrow$ denotes that higher is better.}
  \centering
  \vspace{-1.2em}
   \begin{tabular}{l cccc cccc cccc}
     \toprule 
     \multirow{3}{*}{\textbf{Models}} & \multicolumn{8}{c}{\textbf{Gym-QA \& Diving-QA}} & \multicolumn{4}{c}{\textbf{SPORTU}} \\
     \cmidrule(lr){2-9} \cmidrule(lr){10-13}
     & \multicolumn{4}{c}{\textbf{Action}} & \multicolumn{4}{c}{\textbf{Full}} & \multirow{2}{*}{\textbf{Easy}$\uparrow$} & \multirow{2}{*}{\textbf{Medium}$\uparrow$} & \multirow{2}{*}{\textbf{Hard}$\uparrow$} & \multirow{2}{*}{\textbf{Overall}$\uparrow$} \\
     \cmidrule(lr){2-5}\cmidrule(lr){6-9}
      & \textbf{Event}$\uparrow$ & \textbf{Set}$\uparrow$ & \textbf{Element}$\uparrow$ & \textbf{Overall}$\uparrow$ & \textbf{Easy}$\uparrow$ & \textbf{Medium}$\uparrow$ & \textbf{Hard}$\uparrow$ & \textbf{Overall}$\uparrow$ &  &  &  &   \\
     \midrule
     VideoChatGPT~\cite{maaz2023video} & 35.4 & 22.1 & 8.6 & 22.0 & 34.8 & 29.8 & 8.8 & 25.6 & 37.0 & 36.0 & 22.8 & 34.1 \\
     VideoChat2~\cite{li2024mvbench} & 85.5 & 45.1 & 11.5 & 47.4 & 79.2 & 37.9 & 13.0 & 41.8 & 89.4 & 58.7 & 25.3 & 61.5 \\
     Tarsier~\cite{wang2024tarsier} & 87.9 & 51.3 & 13.3 & 50.8 & 71.1 & 37.9 & 12.3 & 39.6 & 88.1 & 58.4 & 25.2 & 61.0 \\
     \midrule
     LLaVA-Next-Video \cite{zhang2024llavanext-video} & 87.9 & 47.2 & 12.1 & 49.1 & 80.8 & 38.4 & 13.1 & 42.5 & \textbf{92.4} & 59.4 & 30.8 & 63.7 \\
     $+$ VideoTree~\cite{wang2024videotree} & 87.9 & 44.0 & 8.6 & 46.8 & 80.0 & 37.8 & 10.4 & 41.2 & 92.0 & 56.9 & 25.1 & 61.2  \\
     \rowcolor{cyan!10}
     $+$ \textbf{FineQuest (Ours)} & \textbf{88.2} & \textbf{64.6} & \textbf{58.4} & \textbf{70.4} & \textbf{81.0} & \textbf{58.9} & \textbf{49.9} & \textbf{62.1} & \textbf{92.4} & \textbf{71.4} & \textbf{63.7} & \textbf{76.1} \\
     \rowcolor{cyan!10}
     $\Delta\%$ & 0.3 & 36.9 & 382.6 & 43.4 & 0.2 & 53.4 & 280.9 & 46.1 & 0.0 & 20.2 & 106.8 & 19.5 \\
     \midrule
     Video-LLaVA~\cite{lin2023video} & 83.5 & 44.0 & 9.1 & 45.5 & 79.8 & 37.9 & 12.9 & 41.9 & 88.2 & 55.9 & 27.5 & 60.0  \\
     $+$ VideoTree~\cite{wang2024videotree} & 79.6 & 37.5 & 7.4 & 41.5 & 80.0 & 36.7 & 9.8 & 40.5 & 85.1 & 52.6 & 22.3 & 56.5  \\
     \rowcolor{cyan!10}
     $+$ \textbf{FineQuest (Ours)} & \textbf{84.7} & \textbf{59.0} & \textbf{47.2} & \textbf{63.6} & \textbf{80.1} & \textbf{50.0} & \textbf{47.9} & \textbf{57.0} & \textbf{88.7} & \textbf{69.6} & \textbf{59.0} & \textbf{73.2}  \\
     \rowcolor{cyan!10}
     $\Delta\%$ & 1.4 & 34.1 & 418.7 & 39.8 & 0.4 & 31.9 & 271.3 & 36.0 & 0.6 & 24.5 & 114.5 & 22.0  \\
     \hline 
   \end{tabular}
  \label{tab:comparison_sportqa}
  \vspace{-1em}
\end{table*} 
\endgroup

\begingroup
\setlength{\tabcolsep}{2pt}
\begin{table}[t]
\fontsize{8.5}{7.2}\selectfont 
\setlength{\tabcolsep}{1.mm}
  \centering
  \caption{Main results on General VideoQA.}
  \centering
    \vspace{-1.2em}
   \begin{tabular}{l cccc}
     \toprule 
     \textbf{Models} & \textbf{MSVD}$\uparrow$ & \textbf{MSR-VTT}$\uparrow$ & \textbf{TGIF}$\uparrow$ & \textbf{Act.Net}$\uparrow$ \\
     \midrule
     VideoChatGPT~\cite{maaz2023video} & 64.9 & 49.3 & 51.4 & 35.2 \\
     VideoChat2~\cite{li2024mvbench} & 70.0 & 54.1 & - & 49.1  \\
     Tarsier~\cite{wang2024tarsier} & 77.0 & 62.0 & 79.2 & 59.5 \\
     \midrule
     LLaVA-Next-Video \cite{zhang2024llavanext-video} & 73.6 & 57.9 & 71.4 & 53.5 \\
     $+$ VideoTree~\cite{wang2024videotree} & 75.9 & 55.5 & 70.2 & 54.7  \\
     \rowcolor{cyan!10}
     $+$ \textbf{FineQuest (Ours)} & \textbf{77.2} &\textbf{ 58.4} &\textbf{ 72.0} & \textbf{55.3} \\
     \midrule
     Video-LLaVA~\cite{lin2023video} & 71.8 & \textbf{59.0} & 48.4 & 45.3 \\
     $+$ VideoTree~\cite{wang2024videotree} & 73.2 & 56.9 & 48.0 & 47.0  \\
     \rowcolor{cyan!10}
     $+$ \textbf{FineQuest (Ours)} & \textbf{73.8} & \textbf{59.0} & \textbf{49.1} & \textbf{47.8} \\
     \hline 
   \end{tabular}
  \label{tab:comparison_generalqa}
  \vspace{-1.4em}
\end{table} 
\endgroup

\begin{figure}[!t]
    \centering
    \vspace{-0.4em}
   \includegraphics[width=1\linewidth]{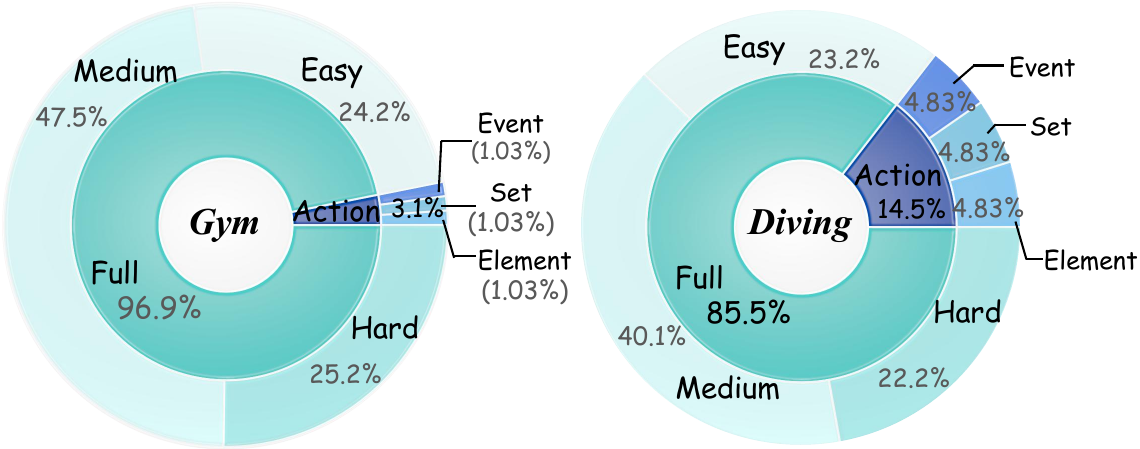}
   \vspace{-2.4em}
    \captionof{figure}{\label{fig:bench}
    Statistics of Gym-QA (\textit{left}) and Diving-QA (\textit{right}). The total number of Gym-QA QA pairs is $27,469$; Diving-QA totaled $1,055$.}
    \vspace{-1.4em}
\end{figure}

\ding{183} \textbf{Relational Scene Level.} To address ambiguities at the instance level, we model relationships within the visual scene graph of $\mathbf{i}_n^S$. For example, the relational triple \{"athlete", "on top", "balance beam"\} is formatted as \textit{"The athlete on top of the balance beam"}. Notably, we further design a \textit{negative} format, \textit{"The athlete \textbf{\textit{not}} on top of the balance beam"}, , to emphasize relational clarity. The visual-to-relation (V2R) similarity is calculated as:
\begin{equation}
    sim_{V2R}^{pos}=\frac{\hat{\mathcal{C}}[j]\cdot\mathbf{r}_{n_{id}}^{pos}}{\left\|\hat{\mathcal{C}}[j]\right\|\left\|\mathbf{r}_{n_{id}}^{pos}\right\|},\quad sim_{V2R}^{neg}=\frac{\hat{\mathcal{C}}[j]\cdot\mathbf{r}_{n_{id}}^{neg}}{\left\|\hat{\mathcal{C}}[j]\right\|\left\|\mathbf{r}_{n_{id}}^{neg}\right\|},
\end{equation}
where the final V2R similarity is defined as: $sim_{V2R}=sim_{V2R}^{pos}-sim_{V2R}^{neg}$. 
By integrating the instance-level and relational-level matching, we select the terminologies and descriptions of the $N_2$ most similar action types from SSGraph $\mathcal{G}_{SS}$ and incorporate them into the reasoning prompt $p_r$ to enrich the knowledge base.

The enriched reasoning prompt $\hat{p_r}=\mathcal{S}_{match}(p_r, \hat{\mathcal{C}}, \mathbf{C}, \mathcal{G}_{SS})$, along with the captions $\mathbf{C}$ and input query $\mathcal{Q}$, is passed to the LLM reasoner $\mathcal{P}_{Reason}(\cdot)$ for zero-shot sports VideoQA:

\vspace{-0.2em}
\begin{equation}
\text{Answer} = \mathcal{P}_{Reason}(\mathcal{E}_V(\mathcal{Q}), \mathcal{E}_L(\hat{\mathcal{C}}, \hat{p_r})).
\end{equation}

\vspace{-0.6em}

\section{Benchmarking High-speed Sports: Gym-QA \& Diving-QA}
\label{sec:bench}

Constructing video understanding datasets, particularly for sports videos, poses significant challenges in terms of video collection and annotation \cite{mangalam2024egoschema, xiao2021next, fang2024mmbench, fu2024video}. Existing benchmarks like Sports-QA \cite{li2024sports} provide limited query depth and diversity, while SPORTU \cite{xia2024sportu} focuses on slow-motion clips. To address these limitations, we introduce two high-quality benchmarks, \textbf{Gym-QA} and \textbf{Diving-QA}, tailored for high-speed sports VideoQA. These benchmarks are constructed using FineGym \cite{shao2020finegym} and FineDiving \cite{xu2022finediving}, with systematically designed QA pairs categorized into three difficulty levels (examples can be found in Fig.\textcolor{magenta}{~\ref{fig:casestudy}}(a)): \textit{Easy}, \textit{Medium}, and \textit{Hard}.
Additionally, we propose a novel \textit{Action Subset}, designed to evaluate the comprehension of intricate visual details in sports, further divided into three coarse-to-fine levels: \textit{Event}, \textit{Set}, and \textit{Element}.
An overview of benchmark statistics is shown in Fig.\textcolor{magenta}{~\ref{fig:bench}}. For detailed construction processes and examples, please refer to \textbf{Appendix}~\S\textcolor{magenta}{C}.




\vspace{-0.6em}
\begingroup
\begin{table}[t]
\fontsize{8.5}{6.2}\selectfont 
\setlength{\tabcolsep}{0.8mm}
\renewcommand{\arraystretch}{1.6}
  \centering
  \caption{Ablation study of components of FineQuest on the full set of Gym-QA and Diving-QA.}
  \centering
  \vspace{-1.2em}
   \begin{tabular}{l cccc}
     \toprule 
     \textbf{Methods} & \textbf{Easy}$\uparrow$ & \textbf{Med.}$\uparrow$ & \textbf{Hard}$\uparrow$ & \textbf{Over.}$\uparrow$ \\
     \midrule
     \rowcolor{cyan!10}
     FineQuest & \textbf{80.1} & \textbf{50.0} & \textbf{47.9} & \textbf{57.0} \\
     \rowcolor{yellow!10}
     w/o $\mathcal{F}_{\text{React}}$ & 78.8 & 50.0 & 48.1 & 56.7 \\
     \rowcolor{yellow!10}
     w/o $\mathcal{F}_{\text{React}},\;\mathcal{S}_{segment}$ & 79.4 & 45.2 & 37.5 & 51.8 \\
     \rowcolor{yellow!10}
     w/o $\mathcal{F}_{\text{React}},\;\mathcal{S}_{segment,\;select}$ & 79.9 & 42.2 & 27.8 & 47.9  \\
     \rowcolor{yellow!10}
     w/o $\mathcal{F}_{\text{React}},\;\mathcal{S}_{segment,\;select,\;match}$ & 79.8 & 37.9 & 12.9 & 41.9   \\
     \bottomrule 
   \end{tabular}
  \label{tab:ablation_vista}
  \vspace{-1.4em}
\end{table} 
\endgroup

\section{Experiment}
\label{sec:experi}

In this section, we empirically investigate the effectiveness of FineQuest in sports VideoQA.

\vspace{-0.8em}
\subsection{Experimental Settings}

\subsubsection*{\textbf{Baselines.}} 
FineQuest is designed to utilize multiple models, \textit{i.e.}, an MLLM as the reactive reasoning agent, a VLM as the captioner, and an LLM as the reasoner. To ensure clarity and fairness in experiments, we select a single MLLM to perform all three roles. Specifically, we apply FineQuest to two different 7B-size MLLMs: Video-LLaVA \cite{lin2023video} and LLaVA-Next-Video \cite{zhang2024llavanext-video}. For Video-LLaVA, it employs LanguageBind \cite{zhu2023languagebind} encoder for visual inputs, and Vicuna-7B v1.5 \cite{vicuna2023} as the LLM backbone. For LLaVA-Next-Video, the visual input is processed through CLIP-ViT-L-336px \cite{radford2021learning} with an MLP projection, with Vicuna as the LLM backbone. While other MLLMs cannot be directly compared due to differences in base models and implementation strategies, we provide results for reference: VideoChat2 \cite{li2024mvbench}, Video-ChatGPT \cite{maaz2023video}, and Tarsier \cite{wang2024tarsier}.

\vspace{-0.44em}
\subsubsection*{\textbf{Evaluations.}} To evaluate the effectiveness of FineQuest, we adopt benchmarks for two aspects: (1) Sports VideoQA: Our proposed Gym-QA and Diving-QA, as well as SPORTU \cite{xia2024sportu}, covering basketball, soccer, ice hockey, tennis, baseball, badminton, and volleyball. (2) General VideoQA: MSVD-QA \cite{xu2017video}, MSR-VTT-QA \cite{xu2017video}, TGIF-QA \cite{jang2017tgif}, and ActivityNet-QA \cite{yu2019activitynet} to ensure that improvements in sports VideoQA do not compromise general QA capabilities. For ablation studies and detailed analysis, FineQuest is primarily evaluated using Video-LLaVA.

\vspace{-0.44em}
\subsubsection*{\textbf{Implementation Details.}} We follow the default settings of Video-LLaVA \cite{lin2023video} and LLaVA-Next-Video \cite{zhang2024llavanext-video} to implement FineQuest. All experiments are conducted on one NVIDIA H100 GPU. Regarding hyperparameters, we set $\alpha_S=0.5$, $\alpha_T=0.3$, and $\alpha_{ST}=0.2$ as default. $N_1$ and $N_2$ are adaptively adjusted based on the input video lengths: $10$ for videos 30 seconds or shorter, and $20$ for videos between 30 and 60 seconds, and so on.

\vspace{-0.6em}
\subsection{Main Results}
\vspace{-0.1em}

Considering the lack of methods specifically focusing on sports VideoQA, we compare FineQuest with the SOTA training-free VideoQA method, VideoTree \cite{wang2024videotree}, across both sports and general VideoQA benchmarks to validate the effectiveness of our approach.

\vspace{-0.4em}
\subsubsection*{\textbf{Sports VideoQA.}}

To benchmark FineQuest, we focused on sports VideoQA and compared it with the previous training-free VideoQA strategy, VideoTree, using MLLM-based models LLaVA-Next-Video \cite{zhang2024llavanext-video} and Video-LLaVA \cite{lin2023video}. As shown in Table\textcolor{magenta}{~\ref{tab:comparison_sportqa}}, we conducted experiments on our proposed Gym-QA and Diving-QA, as well as the SPORTU \cite{xia2024sportu} benchmarks. The results demonstrate that FineQuest significantly enhances the performance of the base models in sports VideoQA tasks.
Notably, as question complexity increases, FineQuest's performance improvements become more pronounced. This highlights the effectiveness of its dual reasoning mode and underscores the importance of benchmarks with diverse difficulty levels for comprehensive evaluation. Additionally, the observed performance drop of VideoTree on Easy and Event-level questions further reveals the limitations of previous training-free paradigms, \textit{i.e.}, clustering-based segmentation and frame-level captioning, in addressing the challenges of sports video analysis.


\vspace{-0.4em}
\subsubsection*{\textbf{General VideoQA.}}

In addition to unlocking the model's powerful capabilities in sports VideoQA, it is equally critical to evaluate its general performance to ensure that improvements in sports-specific tasks do not come at the expense of general QA abilities. To this end, we evaluate FineQuest on four widely-used open-ended general VideoQA benchmarks under a zero-shot setting, as illustrated in Table\textcolor{magenta}{~\ref{tab:comparison_generalqa}}. Notably, FineQuest consistently surpasses VideoTree and achieves performance gains across both base models. This highlights the effectiveness of its structured reasoning process, powered by the reactive reasoning agent, in enhancing the inherent capabilities of the base models.

\begin{figure}[t]
    \centering
    \vspace{-0.2em}
   \includegraphics[width=1\linewidth]{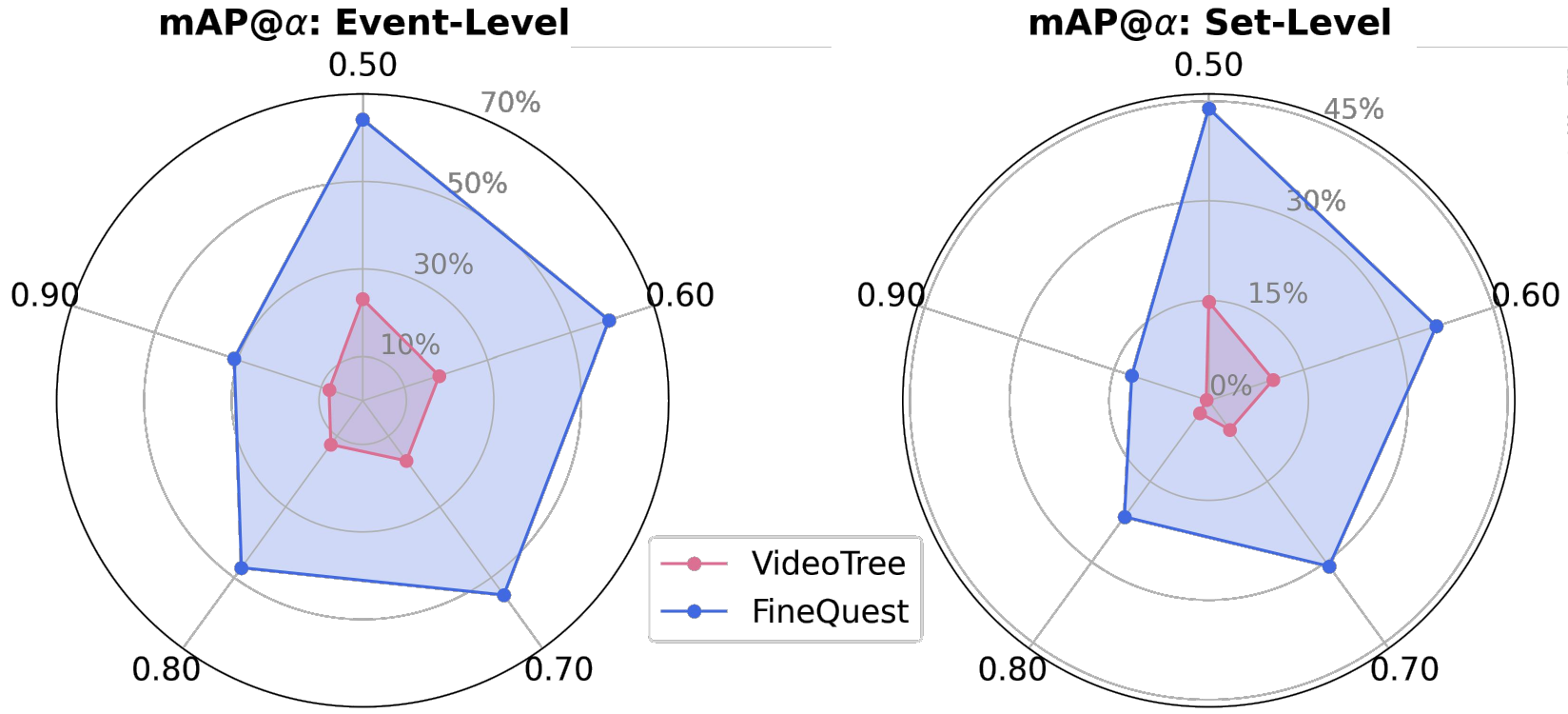}
   \vspace{-2.2em}
    \captionof{figure}{\label{fig:location}
    Temporal localization on FineGym: Event (\textit{left}), Set (\textit{right}). }
    \vspace{-1.4em}
\end{figure}

\vspace{-0.4em}
\subsection{Ablation Studies}
\vspace{-0.1em}

To evaluate the contributions of each module and their combinations, we conduct ablation studies on FineQuest based on Video-LLaVA (Table\textcolor{magenta}{~\ref{tab:ablation_vista}}). The key findings are as follows: \ding{182} \textbf{\textit{Effectiveness of Dual-Reasoning Mode.}} The results clearly demonstrate the importance of the dual-reasoning mode ($\mathcal{F}_{\text{React}}$). When this mode is removed, the performance at the Easy level shows a noticeable decline. This highlights the critical role of FineQuest's reactive reasoning mechanism in enabling dynamic and context-aware reasoning. 
Furthermore, it suggests that using deliberative reasoning for sufficiently simple questions may inadvertently introduce unnecessary information, increasing the risk of hallucinations in such cases.
\ding{183} \textbf{\textit{Importance of Step-by-Step Decomposing Sports Video.}}
The progressive removal of components ($\mathcal{S}_{segment},\;\mathcal{S}_{select}$) further underscores the value of FineQuest’s structured reasoning process. Without these modules, the model struggles to effectively decompose sports videos with complex characteristics (\textit{e.g.}, rapid actions, multi-camera switching) into meaningful segments and extract relevant information step-by-step. This systematic decomposition is essential for managing the fine-grained temporal and spatial dynamics inherent in sports videos, enabling FineQuest to extract meaningful insights and answer questions accurately across varying difficulty levels.
\ding{184} \textbf{\textit{Addressing the Knowledge Gap.}} The $\mathcal{S}_{match}$ module plays a pivotal role in bridging the knowledge gap between the model’s training data or general natural language descriptions and the specialized terminology used in sports. By performing knowledge matching processes, this module aligns general textual descriptions with domain-specific semantics, allowing FineQuest to better interpret and answer questions that require specialized sports knowledge.
Additionally, we conduct hyperparameter ablation studies, as detailed in \textbf{Appendix}~\S\textcolor{magenta}{E}.

\begin{figure}[!t]
    \centering
    \vspace{-0.2em}
   \includegraphics[width=1\linewidth]{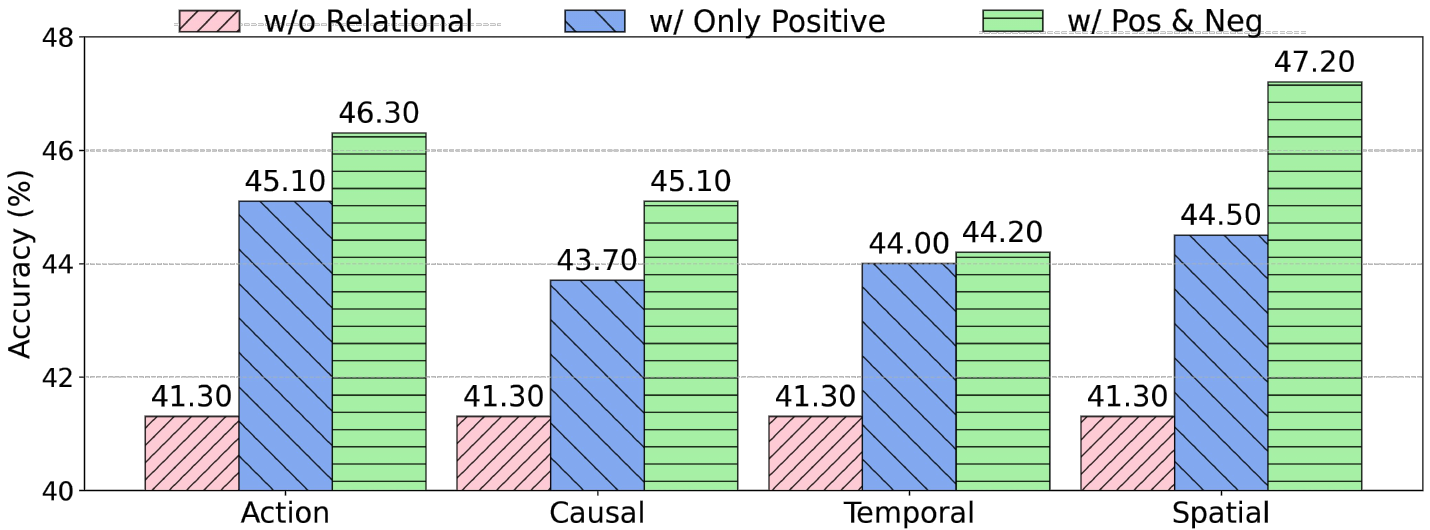}
   \vspace{-2.3em}
    \captionof{figure}{\label{fig:relation}
    Relational ambiguity analysis on the element set of Action within Gym-QA and Diving-QA. "w/o Relational" indicates without relational modeling in Fine-grained Matcher. }
    \vspace{-1.6em}
\end{figure}


\begin{figure*}[!t]
    \centering
    \vspace{-0.2em}
   \includegraphics[width=0.8\linewidth]{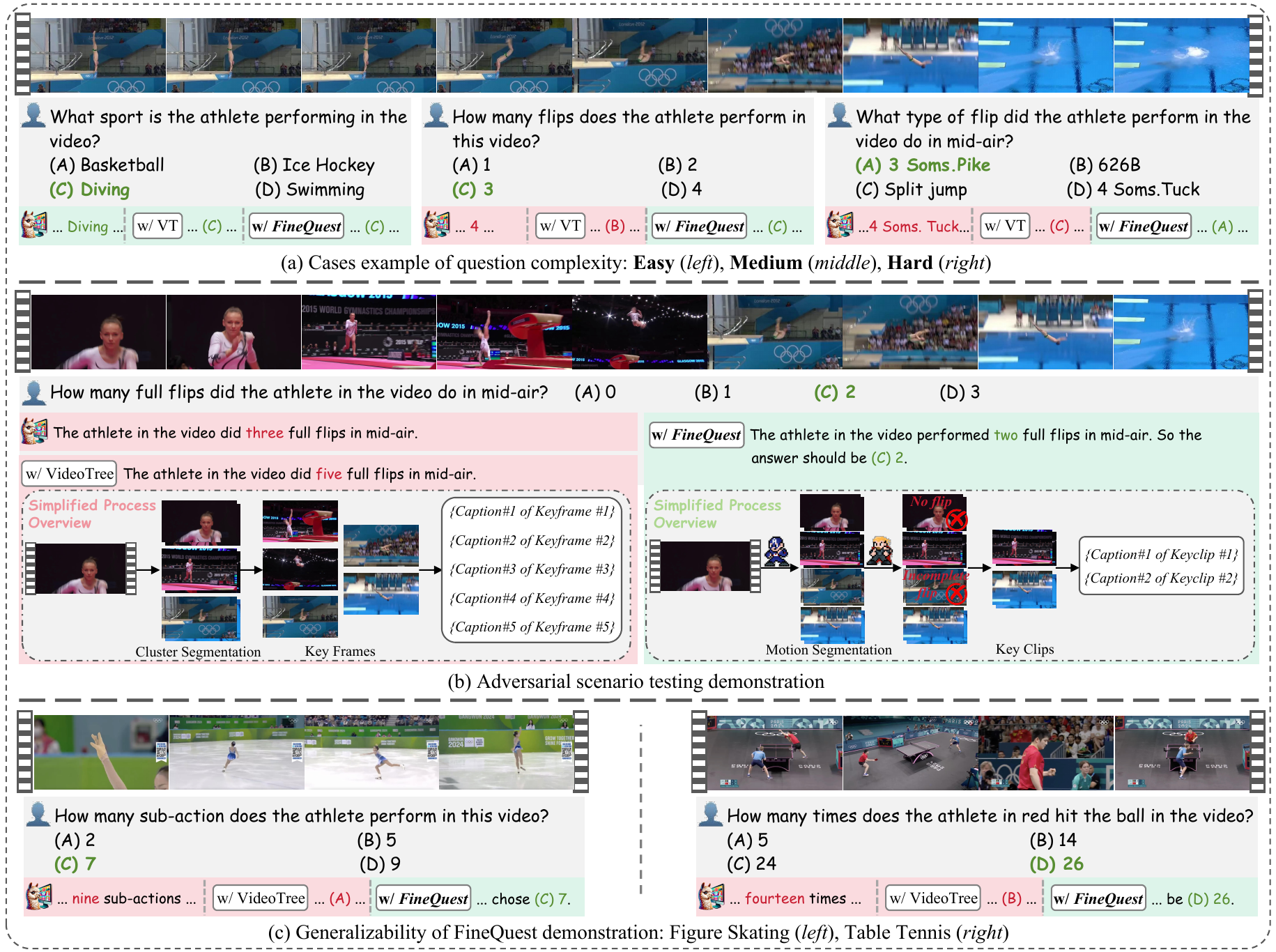}
   \vspace{-1em}
    \captionof{figure}{\label{fig:casestudy}
    Case studies of FineQuest in sports VideoQA, showcasing its effectiveness, robustness, and generalizability. Tests include question complexity (\textit{a}), adversarial scenarios (\textit{b}), and generalizability across diverse sports (\textit{c}), comparing FineQuest with baselines (Video-LLaVA, VideoTree) and ablated versions.}
    \vspace{-0.6em}
\end{figure*}

\vspace{-0.6em}
\section{Analyses and Discussions}

\subsubsection*{\textbf{Effectiveness of Dynamic Motion Segmenter.}}
To empirically demonstrate the effectiveness of the Dynamic Motion Segmenter, we conduct an analysis as illustrated in Figure\textcolor{magenta}{~\ref{fig:location}}. Specifically using $100$ samples ($50$ for event-level, $50$ for set-level) from FineGym \cite{shao2020finegym}. We evaluate temporal action localization with ground truth timestamps, comparing it to VideoTree's clustering-based segmentation. Previous training-free VideoQA methods struggle with the complex and dynamic nature of sports videos, resulting in poor temporal segmentation. In contrast, the Dynamic Motion Segmenter effectively addresses these challenges, achieving significantly higher mean Average Precision (mAP) at both event and set levels. This highlights FineQuest’s ability to adapt to intricate motion patterns and overlapping actions in sports videos, ensuring precise segmentation and enabling more accurate sports VideoQA. We also conduct analysis on the Key Clip Selector, as detailed in \textbf{Appendix}~\S\textcolor{magenta}{E}.

\vspace{-0.5em}
\subsubsection*{\textbf{Relational Ambiguity Analysis.}}
To further analyze the impact of relationship modeling in SSGraph on Fine-grained Matching, we conduct relational ambiguity testing by extending the original scene graph triplets with \textbf{spatial relations} (\textit{e.g.}, \{"athlete", "near", "beam"\}) into \textbf{action relations} (\textit{e.g.}, \{"athlete", "performs", "flip"\}), \textbf{causal relations} (\textit{e.g.}, \{"fall", "causes", "penalty"\}), and \textbf{temporal relations} (\textit{e.g.}, \{"jump", "before", "landing"\}), and testing them individually.
As shown in Figure\textcolor{magenta}{~\ref{fig:relation}}, the introduction of negative relation modeling results in consistent improvements across all four relation types, with spatial relations achieving the most significant gain. This can be attributed to the inherent ambiguity and binary nature of spatial relations (\textit{e.g.}, "near" \textit{vs.} "not near"), making them particularly sensitive to the disambiguation provided by negative relations. FineQuest prioritizes spatial relations for negative relation modeling not only because of their foundational role in scene graph reasoning but also due to the substantial impact of this approach in enhancing spatial alignment and reducing false matches, establishing it as an impactful starting point for relational reasoning.

\begingroup
\begin{table}[h]
\fontsize{8.5}{6}\selectfont 
\setlength{\tabcolsep}{1.4mm}
\renewcommand{\arraystretch}{1.6}
  \centering
  \vspace{-0.6em}
   \begin{tabular}{l|c|c|>{\columncolor{cyan!10}}c}
     \toprule 
     \textbf{Setting} & Video-LLaVA & $+$ VideoTree & $+$ \textbf{FineQuest (Ours)} \\
     \midrule
     \textbf{Original} & 41.0 & 33.0 & \textbf{54.0} \\
     \textbf{Adversarial} & 28.0 & 12.0 & \textbf{46.0} \\
     \bottomrule 
   \end{tabular}
   \caption{Adversarial sceneraio testing.}
  \label{tab:adver}
  \vspace{-2em}
\end{table} 
\endgroup

\vspace{-0.4em}
\subsubsection*{\textbf{Adversarial Scenario Testing.}}
To evaluate FineQuest's robustness, we combined gymnastics and diving video segments to design medium-difficulty questions. For example, Figure\textcolor{magenta}{~\ref{fig:casestudy}}(b) shows a diving segment with incomplete flip actions, increasing motion capture difficulty and introducing scene discrepancies. Previous training-free VideoQA methods struggled with such sports videos, as frame-level captions failed to capture complete action semantics. To assess adversarial discriminative capability, we constructed $50$ adversarial samples and compared them with the original $100$ samples. As shown in Table\textcolor{magenta}{~\ref{tab:adver}}, Video-LLaVA and VideoTree exhibited significant performance declines, while FineQuest retained remarkable performance, highlighting its ability to capture sports actions and enhance adversarial robustness.

\vspace{-0.4em}
\subsubsection*{\textbf{Generalization of FineQuest.}}

As shown in Figure\textcolor{magenta}{~\ref{fig:casestudy}}(c), we further tested FineQuest's generalizability using Olympic data from different sports categories, specifically Figure Skating and Table Tennis. The aim is to assess FineQuest's ability to accurately interpret and analyze sports actions across diverse scenarios. Results showcase FineQuest's capability to adapt and maintain accuracy across diverse sports categories, reinforcing its robustness and versatility in sports VideoQA applications.

\vspace{-0.6em}
\section{Conclusion}
\label{sec:conclu}

In this paper, we introduce \textbf{FineQuest}, the first training-free framework designed for the challenging task of sports VideoQA. FineQuest incorporates a dual-mode reasoning system: 1) \textit{Reactive reasoning} for handling straightforward sports or general queries.
2) \textit{Deliberative reasoning} for addressing more complex queries.
To bridge the gap between general-purpose models and domain-specific sports understanding, we construct \textbf{SSGraph}, a multimodal sports knowledge scene graph covering nine sports. Additionally, to fill the gap in existing high-speed sports QA benchmarks, we introduce two new benchmarks: \textbf{Gym-QA} and \textbf{Diving-QA}. Comprehensive evaluations and analyses show that FineQuest significantly enhances the ability of existing MLLMs to understand sports videos.

\begin{acks}
This work was funded by the National Natural Science Foundation of China (NSFC) under Grant 62306239, and was also supported by National Key Lab of Unmanned Aerial Vehicle Technology under Grant WR202413.
\end{acks}

\bibliographystyle{ACM-Reference-Format}
\balance
\bibliography{sample-base}

\appendix
\newpage

\section{Limitation and Future Work}
\label{app_limitation}

While FineQuest excels at unlocking the sports video understanding capabilities of multimodal large language models, there remains room for exploration in its broader applications, such as sports commentary and acting as a judge or referee in sports scenarios. Additionally, enhancing its domain-specific knowledge base and incorporating more sophisticated reasoning mechanisms could allow FineQuest to better understand and interpret complex sports scenarios, making it a valuable tool for broadcasters and sports analysts alike.

\section{More Details of Dynamic Motion Segmenter}
\label{app_seg}

We provide detailed pseudocode for the Dynamic Motion Segmenter algorithm. This algorithm is a crucial component of our system, designed to accurately segment and analyze motion within sports videos. The pseudocode outlines the step-by-step process used to identify and segment dynamic motions, ensuring precise analysis and understanding of complex sports actions. This detailed breakdown will aid in replicating or adapting the segmenter for various applications within sports video analysis.




\begin{algorithm}[h]
\caption{Dynamic Motion Segmenter}
\begin{algorithmic}[1]
\State \textbf{Require:} $\hat{\mathcal{V}}$: SAM 2 masked video; $win\_size$: initial window size; $z\_range$: range for z-score; $clip\_len$: range for minimum clip length
\vspace{0.4em}
\Function{$\mathcal{S}_{segment}(\cdot)$}{}
    \State $motions \gets \text{GetMotions}(\hat{\mathcal{V}})$ \Comment{Using optical flow}
    \State $proposals \gets [\textcolor{white}{1}]$ \Comment{Initialize proposal list}
    \State $win\_size \gets win\_size$

    \For{$i \in \{win\_size, \ldots, |motions|\}$}
        \State $window \gets \text{motions}[i-win\_size:i]$
        \State $mean \gets \text{mean}(window)$
        \State $std \gets \text{std}(window)$
        \State $z \gets \text{adjust based on variability within } z\_range$
        \State $threshold \gets mean - z \times std$

        \If{$\text{motions}[i] < threshold$ \textbf{and} $i - \text{last\_b} \geq \text{adjusted clip\_len}$}
            \State $proposals\text{.append}(i)$ \Comment{Add new proposal}
        \EndIf
    \EndFor
    \State \Return $proposals$
\EndFunction

\vspace{0.4em}
\Function{$\text{GetMotions}(\hat{\mathcal{V}})$}{}
    \State $motions \gets []$
    \For{$t \in \{1, \ldots, |\hat{\mathcal{V}}|-1\}$}
        \State $frame\_t \gets \hat{\mathcal{V}}[t]$
        \State $frame\_t+1 \gets \hat{\mathcal{V}}[t+1]$
        \State $flow \gets \text{OptFlow}(frame\_t, frame\_t+1)$
        \State $motion \gets \text{AggFlow}(flow)$ \Comment{e.g., magnitude, direction}
        \State $motions\text{.append}(motion)$
    \EndFor
    \State \Return $motions$
\EndFunction
\end{algorithmic}
\label{alg1}
\end{algorithm}


\section{More Details of Gym-QA \& Diving-QA}
\label{app_bench}

\subsubsection*{\textbf{Raw Data Collection.}}
FineGym~\cite{shao2020finegym} and FineDiving~\cite{xu2022finediving} are two high-quality, fine-grained video datasets originally designed for action understanding tasks such as recognition and quality assessment. These datasets feature hierarchical annotations at multiple semantic levels and consist of high-resolution (\textit{e.g.}, 720P and 1080P) official video recordings, ensuring completeness and visual clarity.

To adapt these datasets for VideoQA, we segmented FineGym into $6,031$ event-level clips, each containing $1$ to $32$ element actions, while for FineDiving, we collected videos corresponding to its $31$ action types, each encompassing $3$ to $4$ sub-actions. These adaptations ensure sufficient diversity and complexity, laying a solid foundation for constructing challenging QA tasks.

\vspace{-0.4em}
\subsubsection*{\textbf{Question Answer Generation.}}
Different from existing works (\textit{e.g.}, EgoSchema \cite{mangalam2024egoschema}) that rely on LLMs for QA generation, the subtle visual differences and domain-specific knowledge in sports videos demand a more rigorous approach. We employed a human-centric annotation process with the following key steps: \ding{172} Annotators underwent rigorous training to ensure familiarity with the annotation guidelines. \ding{173} Question generation and answer annotation were performed in separate stages to maintain objectivity, with invalid questions being refined or discarded. \ding{174} Questions were designed to be answerable solely by watching the video, avoiding reliance on general knowledge. 

\vspace{-0.4em}
\subsubsection*{\textbf{Multi-choice Generation.}} To enhance the QA format, open-ended questions were converted to multiple-choice questions. Using GPT-4o(mni) \cite{openai2024hello}, we generated three distractor options for each question, ensuring they were semantically coherent, unique, and distinct from the correct answer. Additionally, we introduced a novel category of \textit{Action Subset}, specifically targeting fine-grained action recognition, which directly evaluates a model's ability to comprehend intricate visual details in sports.

\vspace{-0.4em}
\subsubsection*{\textbf{Query Difficulty Classification.}} For the Full set, we categorized questions into three difficulty levels:
\begin{itemize}[leftmargin=*]
    \item \textbf{Easy}: Requires only a coarse understanding of the video.
    \item \textbf{Medium}: Demands a more fine-grained comprehension of the content.
    \item \textbf{Hard}: Involves domain-specific sports knowledge, challenging the model's ability to understand specialized content.
\end{itemize}
For the Action set, we further classified questions based on the action category into three levels:
\begin{itemize}[leftmargin=*]
    \item \textbf{Event}: Pertains to broader actions or sequences within the sport.
    \item \textbf{Set}: Focuses on more specific segments or routines.
    \item \textbf{Element}: Targets the finest details and individual movements, requiring precise recognition and understanding of intricate visual details.
\end{itemize}
This structured approach ensures a comprehensive assessment of a model's capability to understand and analyze sports videos at varying levels of complexity and detail.

\vspace{-0.4em}
\subsubsection*{\textbf{QA Examination and Filtering.}}
To ensure the quality of Gym-QA and Diving-QA, we implemented a multi-stage filtering pipeline: \ding{182} \textbf{Rule-based filtering}: Pre-configured filters and NLTK tools\footnote{https://github.com/nltk/nltk} were used to clean noise and correct grammatical errors. \ding{183} \textbf{LLM-based blind filtering}: GPT-4o was used to predict answers based solely on the question text, ensuring that questions were grounded in video content. \ding{184} \textbf{Human examination}: Each QA pair, along with its corresponding video, was manually inspected based on our SSGraph to verify accuracy and relevance.

\section{Extended Details of Methodology: Prompts}

In our deployment, we implement an MLLM to act as the reactive reasoning agent. This agent is responsible for executing a structured reasoning process that follows two main stages: \textit{Query Difficulty Analysis} and \textit{Decision Making}. The instruction prompt for this operation is as follows:

\begin{figure}[h]
    \centering
    \vspace{-0.4em}
   \includegraphics[width=1\linewidth]{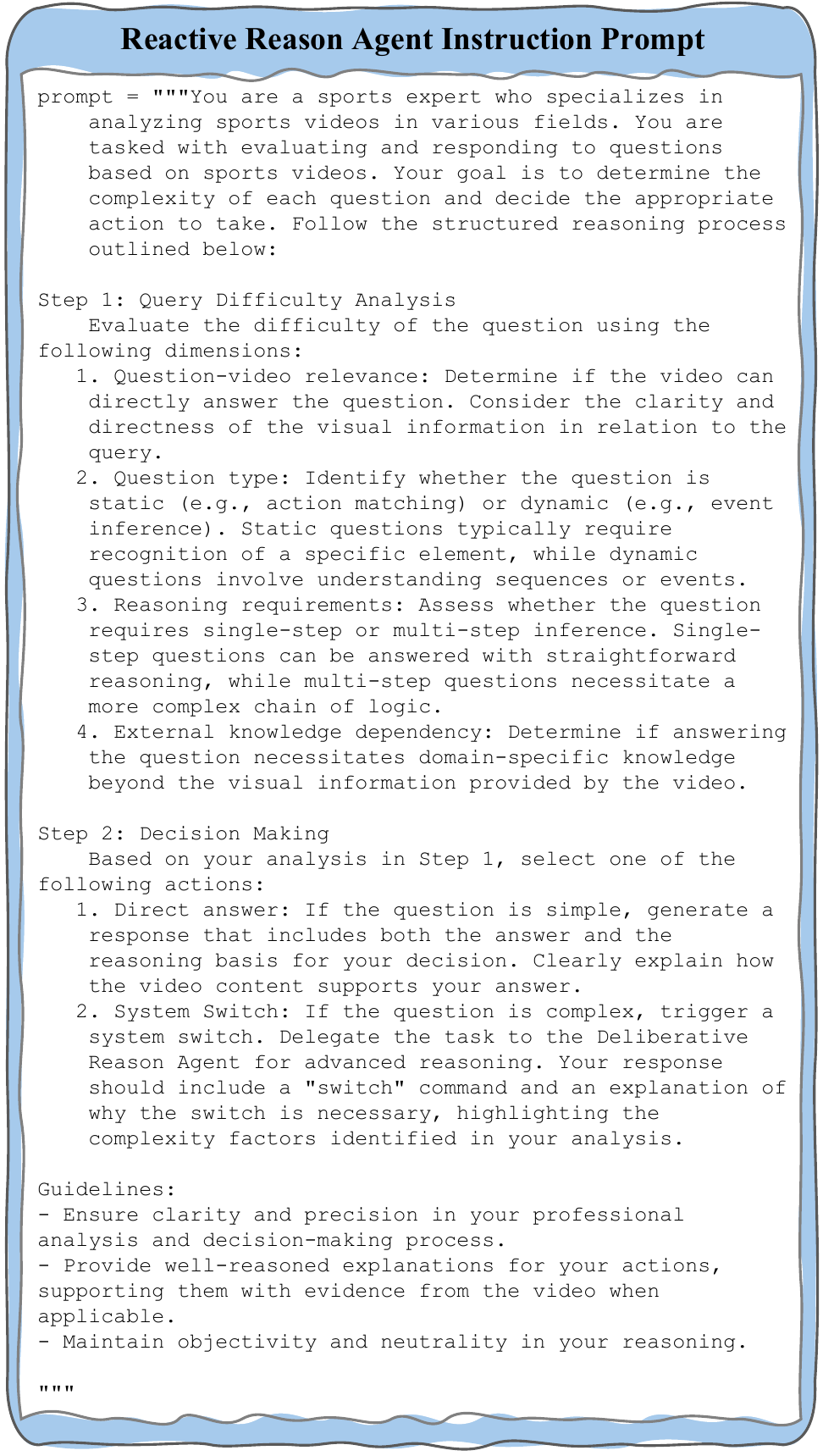}
   \vspace{-2.2em}
    \vspace{-1em}
\end{figure}

\section{More Experimental Analysis}
\label{app_results}

This section presents additional experimental analysis. It includes the hierarchical contrastive decoding of the Key Clip Selector and Hyperparameters of FineQuest.

\subsection{Analysis of Key Clip Selector}

We analyze the impact of the hierarchical contrastive decoding strategy across three dimensions: \ding{192} Spatial, \ding{193} Temporal, and \ding{194} Spatio-Temporal. The results, as shown in Table\textcolor{magenta}{~\ref{tab:ablation_select}}, demonstrate that each dimension of distortion significantly influences the Key Clip Selector's choices. This further validates the effectiveness of employing a multidimensional contrastive strategy when considering videos.

\begingroup
\begin{table}[h]
\fontsize{8.5}{6.2}\selectfont 
\setlength{\tabcolsep}{1.2mm}
\renewcommand{\arraystretch}{1.6}
  \centering
  \centering
  \vspace{-1.2em}
   \begin{tabular}{ccc|cccc}
     \toprule 
     \textbf{Spatial} & \textbf{Temporal} & \textbf{Spa\&Tem} & \textbf{Easy}$\uparrow$ & \textbf{Med.}$\uparrow$ & \textbf{Hard}$\uparrow$ & \textbf{Over.}$\uparrow$ \\
     \midrule
     \rowcolor{yellow!10}
     \ding{55} & \ding{55} & \ding{55} & 78.4 & 49.1 & 32.1 & 50.9 \\
     \rowcolor{yellow!10}
     \ding{51} & \ding{55} & \ding{55}  & 79.2 & 52.9 & 38.9 & 55.1 \\
     \rowcolor{yellow!10}
     \ding{51} & \ding{51} & \ding{55}  & 80.0 & 55.9 & 44.9 & 59.1  \\
     \rowcolor{cyan!10}
     \ding{51} & \ding{51} & \ding{51} & \textbf{81.0} & \textbf{58.9} & \textbf{49.9} & \textbf{62.1} \\
     \bottomrule 
   \end{tabular}
  \caption{Ablation study of the hierarchical contrastive decoding strategy on Full set of Gym-QA and Diving-QA on LLaVA-Next-Video.}
  \label{tab:ablation_select}
  \vspace{-2em}
\end{table} 
\endgroup

The quality of distortion samples significantly influences selection effectiveness. We further explore strategies for spatial and temporal distortions:
\begin{itemize}[leftmargin=*]
    \item \textbf{Spatial}: (i) \textit{Gaussian noise:} Adds random noise to frames. (ii) \textit{CutMix:} Combines parts of different images. (iii) \textit{Blur:} Obscures details and edges. (iv) \textit{Color Jitter:} Alters brightness and color.
    \item \textbf{Temporal}: (i) \textit{Temporal Warping:} Distorts the temporal duration of each frame (ii) \textit{All Shuffle:} Randomly shuffles all video frames. (iii) \textit{Local Shuffle:} Randomly shuffles partial video frames. (iv) \textit{Reverse:} Reverses the temporal order.
\end{itemize}

\begin{figure}[h]
    \centering
    \vspace{-1.4em}
   \includegraphics[width=1\linewidth]{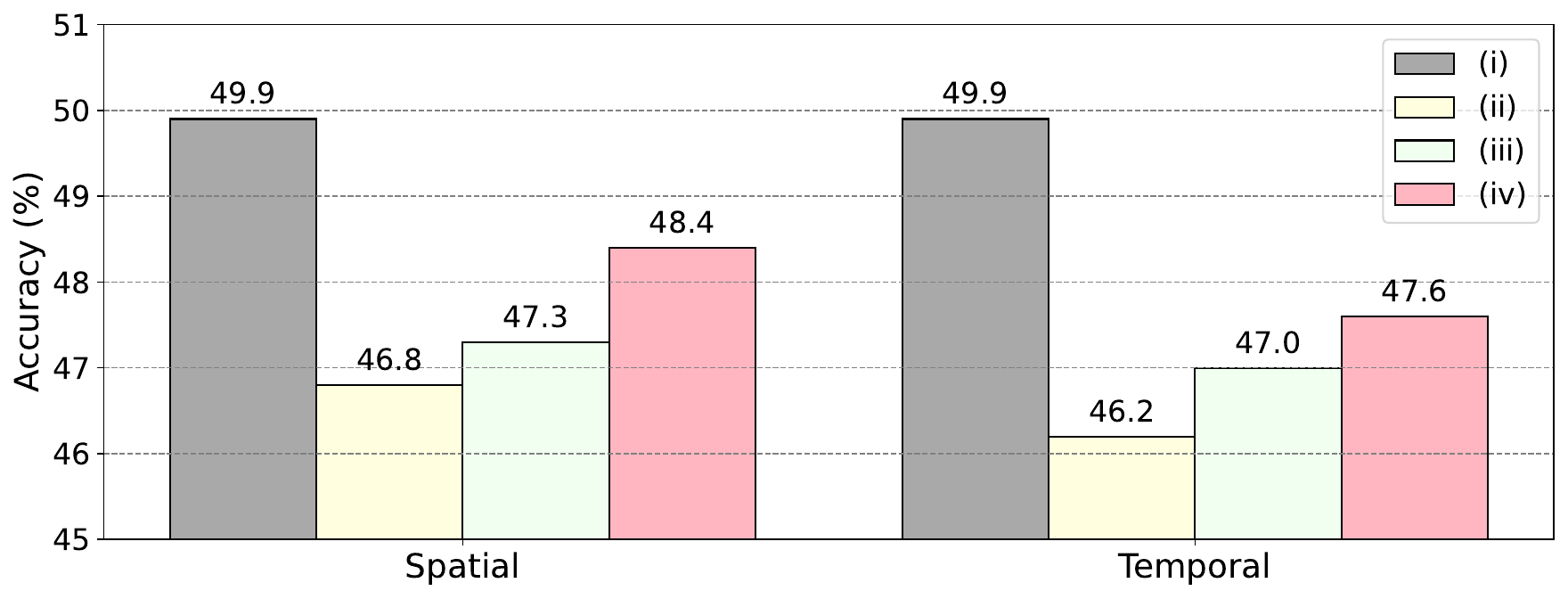}
   \vspace{-2.6em}
    \captionof{figure}{\label{fig:select}
    Ablation study of distortion variants on Hard level of Gym-QA and Diving-QA.}
    \vspace{-1em}
\end{figure}

As demonstrated in Figure\textcolor{magenta}{~\ref{fig:select}}, we observe the following performance trends: In the \textbf{\textit{spatial}} dimension, Gaussian noise achieves the highest accuracy, suggesting that moderate noise can help the model maintain high performance by simulating real-world uncertainties. In contrast, CutMix performs the worst, likely due to its alteration of the spatial structure, which makes it difficult for the model to recognize objects effectively. Blur and color jitter strategies, on the other hand, show relatively high accuracy, indicating the model's robustness to blurring and color variations. In terms of \textbf{\textit{temporal}} distortions, temporal warping stands out as the most effective strategy, highlighting the importance of maintaining frame order, even with slight temporal distortions, to capture motion dynamics accurately. Conversely, the all-shuffle strategy performs the worst, as completely disrupting the sequence significantly impacts the model's understanding of temporal information. Local shuffle and reverse strategies have a moderate impact, suggesting that the model can adapt to some degree of temporal changes. These observations indicate that while moderate perturbations can enhance model generalization, excessive distortions tend to degrade performance.

\subsection{Analysis of Hyperparameters}

\begin{figure}[h]
    \centering
    \vspace{-0.4em}
   \includegraphics[width=1\linewidth]{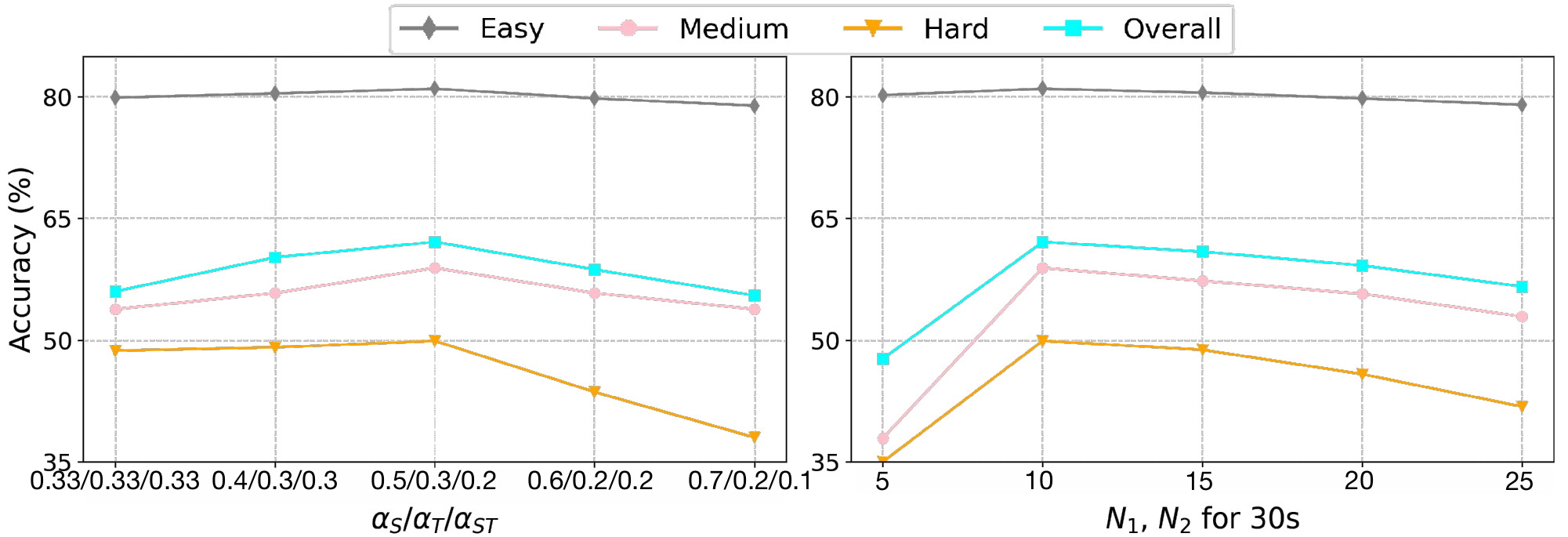}
   \vspace{-2.2em}
    \captionof{figure}{\label{fig:hyper}
    Ablation study of hyperparameters.}
    \vspace{-1em}
\end{figure}

We analyze the impact of two hyperparameter sets of FineQuest performance in Figure\textcolor{magenta}{~\ref{fig:hyper}}:
\ding{192} \textit{Weights of hierarchical contrastive decoding ($\alpha_S,\;\alpha_T,\;\alpha_{ST}$):} It is evident that the configuration with a spatial weight of $0.5$, a temporal weight of $0.3$, and a spatio-temporal weight of $0.2$ yields the best overall performance. This suggests that placing greater emphasis on spatial features, while maintaining a balanced consideration of temporal and spatio-temporal elements, optimizes the model's ability to decode effectively in a hierarchical manner.
\ding{193} \textit{Number of clips sampling ($N_1,\;N_2$):} The analysis shows that setting $N=10$ achieves the highest accuracy across different difficulty levels (Easy, Medium, and Hard). This indicates that a moderate threshold allows for an optimal balance between capturing sufficient information and avoiding excessive noise, thereby enhancing the model's ability to select the most relevant clips for accurate assessment. These findings highlight the importance of carefully tuning both the weight distribution in hierarchical contrastive decoding and the clip selection threshold to maximize model performance.

\end{document}